\pdfoutput=1
\documentclass[sigconf,nonacm]{acmart}
\acmConference[HRI '26]{Proceedings of the 21st ACM/IEEE International Conference on Human-Robot Interaction}{March 16--19, 2026}{Edinburgh, Scotland, UK}
\acmBooktitle{Proceedings of the 21st ACM/IEEE International Conference on Human-Robot Interaction (HRI '26), March 16--19, 2026, Edinburgh, Scotland, UK}

\acmDOI{10.1145/3757279.3785585}
\acmISBN{979-8-4007-2128-1/2026/03}

\usepackage{amsmath}
\usepackage{graphicx}
\usepackage{xcolor}
\usepackage{gensymb}

\ccsdesc[500]{Human-centered computing~Pointing}
\ccsdesc[500]{Human-centered computing~Text input}
\ccsdesc[500]{Human-centered computing~Gestural input}
\ccsdesc[300]{Computer systems organization~Robotic components}
\ccsdesc[500]{Computer systems organization~Real-time system architecture}

\keywords{HRI, POMDP, multimodal fusion, gesture, language}

\title{LEGS-POMDP: Language and Gesture-Guided Object Search in Partially Observable Environments}

\author{Ivy Xiao He}
\orcid{0009-0001-0141-1915}
\affiliation{%
  \institution{Brown University}
  \city{Providence}
  \country{USA}
}
\email{xiao_he@brown.edu}

\author{Stefanie Tellex}
\orcid{0000-0002-2905-4075}
\affiliation{%
  \institution{Brown University}
  \city{Providence}
  \country{USA}
}
\email{stefie10@cs.brown.edu}

\author{Jason Xinyu Liu}
\orcid{0000-0001-7732-3666}
\affiliation{%
  \institution{Brown University}
  \city{Providence}
  \country{USA}
}
\email{xinyu_liu@brown.edu}

\begin{document}
\begin{abstract}
To assist humans in open-world environments, robots must interpret ambiguous instructions to locate desired objects.
Foundation model-based approaches excel at multimodal grounding, but they lack a principled mechanism for modeling uncertainty in long-horizon tasks. In contrast, Partially Observable Markov Decision Processes (POMDPs) provide a systematic framework for planning under uncertainty but are often limited in supported modalities and rely on restrictive environment assumptions.
We introduce \underline{L}anguag\underline{E} and \underline{G}e\underline{S}ture-Guided Object Search in Partially Observable Environments (\textbf{LEGS-POMDP}), a modular POMDP system that integrates language, gesture, and visual observations for open-world object search.
Unlike prior work, LEGS-POMDP explicitly models two sources of partial observability: uncertainty over the target object’s identity and its spatial location.
In simulation, multimodal fusion significantly outperforms unimodal baselines, achieving an average success rate of $89\%\pm7\%$ across challenging environments and object categories.
Finally, we demonstrate the full system on a quadruped mobile manipulator, where real-world experiments qualitatively validate robust multimodal perception and uncertainty reduction under ambiguous instructions.
\end{abstract}

\maketitle

\section{Introduction}

\begin{figure}
  \centering
  \includegraphics[width=\linewidth]{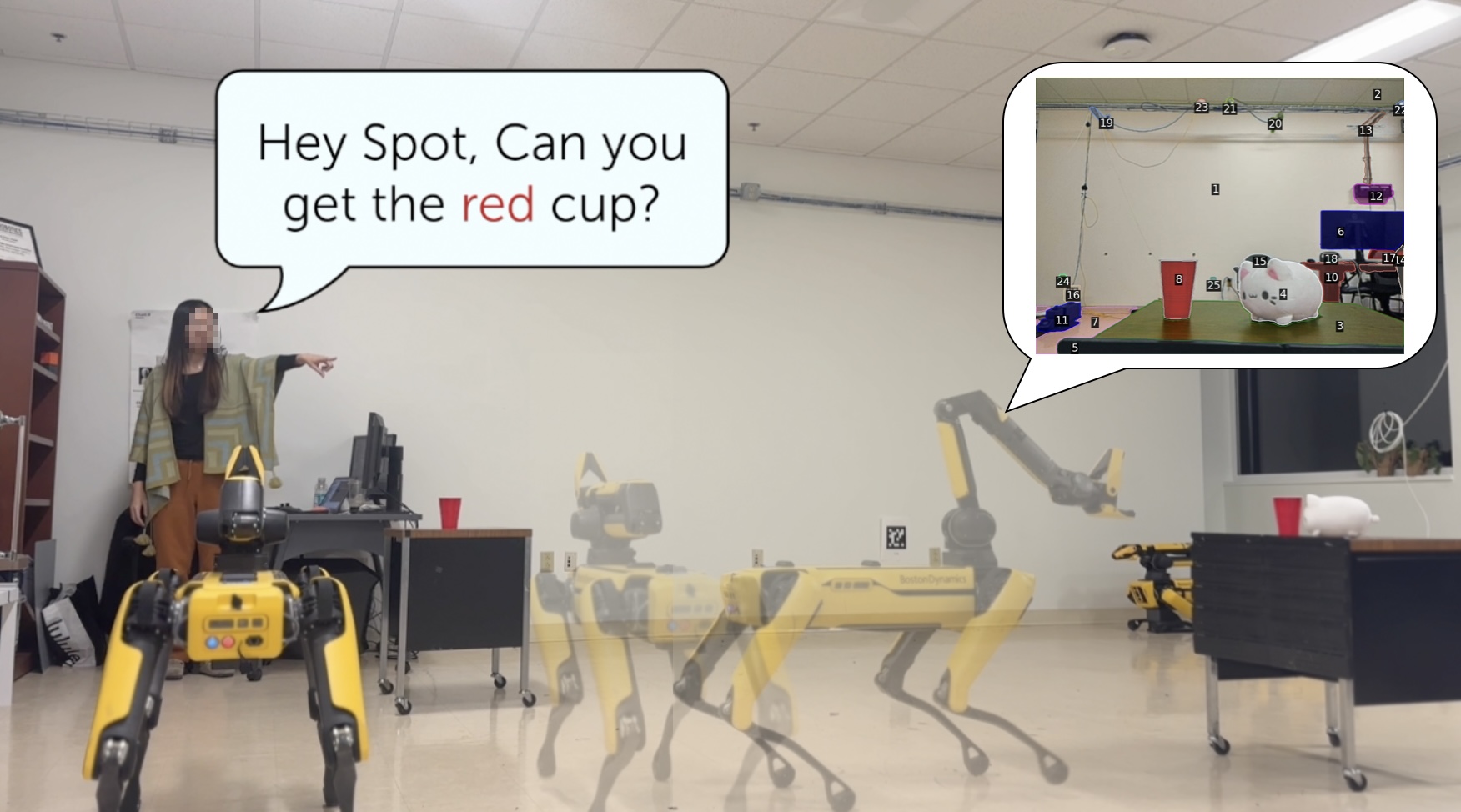}
  \caption{Multimodal fusion with belief updates disambiguates human instructions and identifies the intended object among multiple candidates.}
  \Description{Given a spoken instruction and a pointing gesture, the robot disambiguates the target object among multiple candidates and retrieves the correct one.}
  \label{fig:teaser}
\end{figure}

To assist humans in unstructured open-world environments, robots must accurately understand and act upon ambiguous instructions to find target objects.
The human-instructed object search problem requires robots to both identify which object is being referred to and determine where it is located, under uncertainty arising from underspecified language, imprecise gestures, and noisy perception. As illustrated in Figure~\ref{fig:teaser}, language alone may be vague, gestures may indicate regions containing multiple candidates, and sensor noise further compounds ambiguity.
A key observation is that different modalities are often complementary: gestures can disambiguate vague language, while language can clarify imprecise gestures. Although humans naturally combine language and gesture during communication, enabling robots to robustly interpret such multimodal cues in partially observable environments remains challenging. To achieve multimodal referring expression understanding~\cite{mao2016generation} in these settings, robots must jointly reason over uncertainty in language, gesture, and visual perception.

Existing work on human-instructed object search largely falls into two categories, each with limitations in open-world scenarios.
Foundation-model-based methods that ground multimodal input and directly generate actions~\cite{lin_gesture-informed_2023, mane2025ges3vig} often lack explicit uncertainty modeling and long-horizon sequential decision-making, and they provide limited formal guarantees and interpretability.
Moreover, collecting large-scale datasets of natural referential gestures for fine-tuning remains difficult~\citep{doi:10.1126/scirobotics.adf6991, filipek2025empirical, mane2025ges3vig, caesar_islam_2022}.
In contrast, POMDPs explicitly model uncertainty for sequential decision-making.
However, prior POMDP-based object search has largely focused on tabletop settings~\citep{zhang_invigorate_2024}, relied on language alone, or made restrictive environment assumptions~\citep{whitney_reducing_2017, zheng_system_2023}.



To address these limitations, we introduce \underline{L}anguag\underline{E} and \underline{G}e\underline{S}ture-Guided Object Search in Partially Observable Environments (\textbf{LEGS-POMDP}), a modular POMDP framework that integrates language, gesture, and visual observations for open-world object search. LEGS-POMDP explicitly models two sources of partial observability: uncertainty over the human’s intent (target object identity) and uncertainty over the environment (target object location). By maintaining joint beliefs over object identity and location, the robot can reason over both instruction-level and environment-level ambiguity and produce explainable decision-making behavior.

Our multimodal observation model leverages state-of-the-art language, gesture, and visual perception modules to represent each modality as a likelihood function over candidate objects, which are fused in log-space to form a joint observation distribution. The modular design enables flexible replacement or upgrading of individual perception components, while preserving principled Bayesian belief updates and interpretability that are difficult to achieve with end-to-end approaches.


We evaluate LEGS-POMDP through both modular reference understanding benchmarks and full-system decision-making experiments to assess multimodal object search under uncertainty.
Specifically, our evaluation includes: (i) gesture grounding with five different pointing representations; (ii) visual grounding via Set-of-Marks prompting~\cite{yang_set--mark_2023} and Grounding DINO~\cite{liu2024grounding}; (iii) visual sensor models with different fan-shaped configurations to test modularity and parameterization; (iv) full-system evaluation in simulated environments with varying levels of complexity and instruction ambiguity; and (v) real-robot evaluation on a quadruped mobile manipulator. 

This paper makes three key contributions:
(1) We formulate human-instructed object search as a POMDP with two sources of partial observability, explicitly modeling uncertainty over target object identity and spatial location.
(2) We propose a modular multimodal observation model that integrates language, gesture, and visual perception as probabilistic likelihoods within a principled Bayesian belief update.
(3) We evaluate the proposed framework through extensive simulation experiments under varying levels of instruction ambiguity, and qualitatively validate uncertainty reduction on a real quadruped mobile manipulator.

\section{Related Work}

Human-instructed object search is challenging because the robot must handle state uncertainty, perceptual noise, and reference ambiguity.\citep{yokoyama2024vlfm} Prior research has followed two main paradigms: end-to-end learning based methods and modular approaches.
End-to-end methods map multimodal sensor inputs directly to actions, learning semantic priors that support goal-directed exploration and generalization~\citep{doi:10.1126/scirobotics.adf6991, lin2020multi, padalkar2024rtx,chaplot2020objectgoalnavigationusing}. While some approaches introduce intermediate structure, such as visual state abstractions or topological representations, end-to-end learning remains highly data-intensive and often requires large-scale training\citep{garg2024robohopsegmentbasedtopologicalmap, hughes2024foundations, padalkar2024rtx}. 

Modular approaches, on the other hand, decompose the task into perception, semantic grounding, and planning components.\citep{geng2023gapartnet, wang2024sim2real}
This structure facilitates engineering, preserves explainability, and enables changeable modules. Recent modular learning approaches further replace hand-designed components with learned ones while retaining the overall pipeline, thereby combining the data efficiency and sim-to-real transfer benefits of modularity with the representational power of learning.\citep{fu2024mobilealohalearningbimanual, chi2023diffusionpolicy, geng2023gapartnet} 
Building on this paradigm, researchers have extensively explored gesture grounding, language grounding, and multimodal fusion, as well as decision-theoretic frameworks for planning under uncertainty. 


\textbf{\textit{Language Grounding:}}
Grounding natural language commands has long been a central challenge for robots \citep{tellex_robots_nodate, cohen2024ground}. 
Previous works have grounded human instructions to a formal representation and latent space for planning and control~\cite{hsiung2022generalizing, liang2022cap, liu2023lang2ltl, liu2024lang2ltl2, octo2023, brohan2023rt1, zitkovich2023rt2, padalkar2024rtx, kim2024openvla, black2024pi0}.
Interactive approaches such as INGRESS \citep{shridhar_interactive_2018} and attribute-guided POMDP frameworks \citep{yang_interactive_2022} show that asking clarification questions can mitigate linguistic ambiguity. Other work has embedded language directly into observation models \citep{nguyen_language-conditioned_2023}, or leveraged social feedback to reduce misinterpretations in object fetching tasks \citep{whitney_reducing_2017}. Spatial language understanding in large-scale environments further highlights how ambiguity grows when many candidate objects are present \citep{zheng_spatial_2021}. Our framework extends this literature by jointly modeling language ambiguity alongside gesture uncertainty in a unified probabilistic planning framework.

\textbf{\textit{Gesture Grounding:}}
Pointing is a natural and frequent modality in human–robot interaction, often used to resolve referential ambiguity. Early work formalized pointing with geometric models such as the pointing cone \citep{gibet_deixis_2006, nickel2003pointing, nickel2007visual}, while later studies analyzed human pointing behaviors in household settings \citep{electronics14122346, filipek2025empirical}, highlighting the prevalence of ambiguity. More recent approaches incorporate skeletal vectors (eye–wrist, shoulder–wrist) to probabilistically model gesture likelihoods \citep{pelgrim_find_2024}, and integrate pointing into situated language understanding \citep{perlmutter_situated_2016}. With the rise of large models, systems like GIRAF \citep{lin_gesture-informed_2023} and GestLLM \citep{kobzarev_gestllm_2025} emphasize the semantic and contextual nature of gesture interpretation, while visual prompting methods leverage pointing for downstream VQA \citep{tanada_pointing_nodate}. Despite this progress, gesture interpretation remains inherently uncertain due to human variability and sensor noise. Our work builds on this line by explicitly modeling gesture as a probabilistic observation within a POMDP, rather than as a deterministic cue, allowing the robot to reason probabilistically about human intent.

\textbf{\textit{Multimodal Fusion:}} Many systems integrate gesture and language at the perception level to improve disambiguation. 
 while visual prompting methods use pointing for VQA ~\citep{tanada_pointing_nodate, mani_point_2022}. 
These efforts demonstrate that multimodal cues can significantly reduce referential ambiguity; however, fusion is typically confined to the perceptual level and is not connected to downstream decision-making.
Systems such as GIRAF~\citep{lin_gesture-informed_2023} and This\&That~\citep{wang_thisthat_2024} show promising integration of multimodal instructions with robot execution, but their reliance on tabletop domains highlights the need for frameworks that extend to unstructured, large-scale environments.
Complementary work has introduced benchmarks targeting perception challenges, such as open-vocabulary segmentation~\citep{zou_segment_2023, li_semantic-sam_2023}, or multimodal disambiguation datasets that probe gesture and language integration~\citep{chen2021yourefitembodiedreferenceunderstanding, caesar_islam_2022,mane2025ges3vig,nakamura2023deepointvisualpointingrecognition}. These directions highlight the importance of multimodal fusion for instruction following, while also pointing to the need for frameworks that connect multimodal fusion with downstream tasks. 

\textbf{\textit{POMDP \& Uncertainty in HRI: }}
POMDPs provide a principled framework for decision-making under uncertainty, with online solvers such as POMCP~\citep{silver_monte-carlo_nodate} enabling scalability. They have been applied to multi-object search~\citep{wandzel_multi-object_2019}, where language serves as a prior over candidate states, and extended in systems such as GenMOS~\citep{zheng_system_2023} and INVIGORATE~\citep{zhang_invigorate_2024} that integrate visual grounding and interactive dialogue. Yet prior POMDP-based work has largely focused on language cues and has rarely incorporated gesture as a probabilistic observation\citep{wandzel_multi-object_2019, zhang_invigorate_2024, whitney_reducing_2017}. Our work advances this trajectory by explicitly modeling two layers of uncertainty, i.e., human intent and environment state, and showing that multimodal fusion of gesture and language improves efficiency in object search.

\section{Technical Approach}
\begin{figure}[t]
  \centering
  \includegraphics[width=\linewidth]{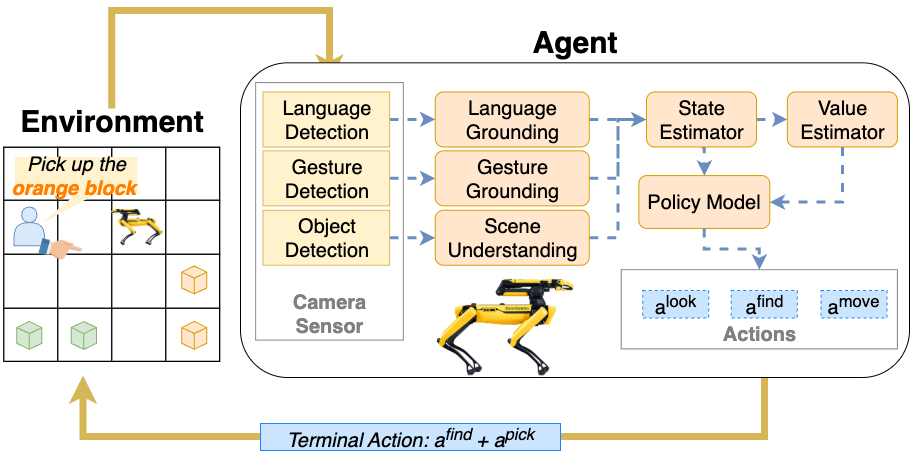}
  \caption{system diagram.}
  \label{fig:system_diagram}
  \Description{system diagram}
\end{figure}


Our system aims to address the language- and gesture-conditioned object search problem, 
where the robot must interpret uncertain human instructions while exploring a partially observed environment. This requires solving the subproblems of: (i) representing robot's uncertainty and the hidden state of the world, (ii) integrating multimodal human instructions, and (iii) planning efficiently under ambiguous instruction and robot uncertainty. This problem can be naturally modeled as a Partially Observable Markov Decision Process (POMDP), 
since the target object’s location is hidden, human inputs are noisy and ambiguous, 
and the robot must plan sequences of actions under uncertainty. Figure~\ref{fig:system_diagram} illustrates the overall architecture of our framework, showing how language, gesture, and vision modules are integrated with the POMDP-based planner.

\subsection{POMDP Formulation}

In order to represent both the robot’s knowledge of the world and its uncertainty about the hidden target, we define the task as a POMDP tuple $(S, A, T, O, Z, R, \gamma)$ as follows:

\textit{State Space:} Each state $s \in \mathcal{S}$ is defined as $s = (s_r, s_o)$, where $s_r = (x,y,\theta)$ is the robot pose, and $s_o$ denotes the latent target location. The obstacle map is known and fixed. We use an object-independent state representation, where objects are labeled as target or distractor based on human intent, rather than category, allowing the framework to focus on uncertainty reasoning rather than object taxonomy.

\textit{Action Space:} The action space is discrete. The agent choose from three classes of actions: movement actions $a_{\texttt{move}}$, an observation-gathering action $a_{\texttt{look}}$, and a termination action $a_{\texttt{find}}$. Move actions use four deterministic motion primitives (forward, backward, turn-left, turn-right) defined in the robot’s relative frame. This abstraction allows flexible combinations of primitives without changing the POMDP formulation.

\textit{Transition Model:} The transition model $T(s' \mid s,a)$ updates the robot pose deterministically. For movement actions,
$T(s' \mid s, a_{move}) = 1$, where robot pose is updated based on the executed primitive. The look action is designed solely to acquire additional multimodal observations for belief update. The target-object location $s_o$ is static and remains unchanged across transitions. 

\textit{Observation Space:} Observations $o = (o_v, o_g, o_l)$ contain multimodal signals from vision, gesture and language instructions. 

\textit{Observation Model:} $Z(o \mid s)$ defines the likelihood of multimodal signals conditioned on the hidden state, with vision, gesture, and language terms fused by a weighted log-likelihood. 

\textit{Reward Model:} Reward $R(s,a)$ assigns a positive reward for a correct $a_\texttt{find}$, a small negative cost for $a_\texttt{move}$ and $a_\texttt{look}$ to encourages efficient exploration. This sparse structure ensures that the planner prioritizes correct target finding, while step costs discourage exhaustive exploration. Discount factor $\gamma \in (0,1)$ balances immediate and future rewards.

\subsection{Multimodal Observation Model}
In order to integrate human instructions with perceptual signals, we design an observation model that fuses three modalities: vision, language, and gesture. Unlike end-to-end models, this modular observation formulation is less data-hungry and provides interpretable likelihoods for each modality, enabling explicit reasoning about uncertainty and explainability in downstream planning.
Each modality is modeled by its likelihood $P(o_m \mid s)$, representing the probability of observing signal $o_m$ given a hypothesized state $s$. 
The contribution of each modality can be controlled by modality-specific weights $(w_v, w_g, w_l)$. 
The modality-specific likelihoods are combined by the fusion model:
\begin{equation}
\log Z(o\mid s) =
w_v \log P_v(o_v \mid s) +
w_g \log P_g(o_g \mid s) +
w_l \log P_l(o_l \mid s),
\label{eq:loglikelihood_fusion}
\end{equation}
Our observation model can be formulated as:
\begin{equation}
Z(o \mid s) \;\propto\; \prod_{m \in \{v,g,l\}} P_m(o_m \mid s)^{w_m}.
\end{equation}
This formulation naturally integrates with Bayesian belief updates as shown in Eq.\ref{eq:bayesian_update}, where multimodal likelihoods provide evidence to reweight the posterior over hidden states.
\begin{equation}
b^\prime(s^\prime) \propto Z(o \mid s^\prime) \sum_{s \in S} T(s^\prime \mid s, a)\, b(s).
\label{eq:bayesian_update}
\end{equation}

\paragraph{Visual Observation.}
Camera sensors provide incomplete and noisy detections of objects due to limited field of view and distance-dependent accuracy.  
To approximate this uncertainty, segmentation outputs are treated as candidate object detections, and the camera is modeled as a decaying fan-shaped sensor, similar to \cite{wandzel_multi-object_2019}. 
The likelihood of correctly detecting a target at location $(x,y)$ is defined by Gaussian decay in both angular deviation and range:
\begin{equation}
P_v(o_v=1 \mid s) \;\propto\;
\exp\left(-\frac{\theta_{\text{diff}}^2}{2\sigma_\theta^2}\right)
\cdot
\exp\left(-\frac{(r-r_0)^2}{2\sigma_r^2}\right),  
\label{eq:vision_model}
\end{equation}
 where $\theta_{\text{diff}}$ is the angular difference between the camera’s central axis and the object direction, $r$ is the distance from the robot to the object, and $r_0$ is the nominal detection range. 
This formulation captures the intuition that detections are most reliable when the object is centered in view and within a favorable distance, and less reliable otherwise.

\paragraph{Language Observation.}
Natural language instructions describe the target object but can be ambiguous and error-prone after automatic speech recognition. For example, the same object may be referred to as “cup,” or “mug,” and transcription errors may further increase uncertainty. 
Thus, the challenge is to map an utterance $u_l$ into a probabilistic signal that reflects how well each candidate state location $s_o$ matches the instruction. 
We model this through a similarity function $\kappa(s_o; u_l) \in [0,1]$, which measures how well the hypothesized target at $s_o$ aligns with the given instruction. 
This score is then converted into a likelihood by interpolating between modality-specific false- and true-positive rates:
\begin{equation}
P_l(o_l \mid s)
=\epsilon_l^- + (\epsilon_l^+ - \epsilon_l^-)\,\kappa(s_o;u_l),
\qquad
0<\epsilon_l^-<\epsilon_l^+<1.
\label{eq:language model}
\end{equation}
$\epsilon_l^-$ is the minimum likelihood assigned to irrelevant objects (false positives), and $\epsilon_l^+$ is the maximum likelihood for a perfectly matching description (true positives). 
This formulation captures graded confidence rather than a binary match, allowing the belief update to weigh language input proportionally to its semantic specificity.

\paragraph{Gesture Observation.}
Pointing gestures provide a strong cue about the intended target, but they are inherently uncertain due to human variability and perceptual noise. Humans adopt different pointing strategies: sometimes extending the whole arm, 
sometimes aligning the gaze with the hand, and sometimes raising only the forearm casually. 

To account for this variability, we define the pointing direction dynamically as the mean vector of multiple anatomical cues: eye-to-wrist, shoulder-to-wrist, and elbow-to-wrist. 
The gesture is then represented as a spatial cone with the wrist as the origin and this averaged vector as the central axis. The opening angle of the cone captures the spread of the three vectors.
The likelihood of the target being at location $(x,y)$ is then defined as
\begin{equation}
P_g(o_g \mid s) = \exp\!\left(-\tfrac{\theta_{\text{diff}}^2}{2\sigma_{g}^2}\right),
\label{eq:gesture_likelihood}
\end{equation}
where $\theta_{\text{diff}}$ is the angular deviation between the central pointing vector 
and the vector from the wrist to $(x,y)$, and $\sigma_{g}$ determines the spread of the cone. 
This formulation captures the intuition that states closer to the pointing direction are more likely, 
while off-cone states receive exponentially lower likelihood.

For planning, we employ Partially Observable UCT (PO-UCT) as the solver in both simulation and real-world tests. 
PO-UCT is a Monte Carlo tree search algorithm that balances exploration and exploitation by simulating trajectories 
from the current belief. 
Although not novel in itself, PO-UCT provides a strong, well-established baseline that integrates naturally 
with our multimodal observation models and supports deployment on the Boston Dynamics Spot robot. 
This unified choice allows us to attribute performance differences to perception and grounding quality, 
rather than planning artifacts.

To enable systematic evaluation, we first implement explicit probabilistic observation models in simulation, 
including a fan-shaped vision sensor, a gesture cone model, and a language similarity mapping. In real-robot experiments, however, the robot directly consumes outputs from perception pipelines: 
skeleton tracking for gesture estimation (via MediaPipe), 
a Set-of-Marks (SoM) grounding module combining SAM2 segmentation with GPT-4o reasoning for language input, 
and onboard object detection from the Spot camera system. This design ensures that the POMDP framework accommodates both analytic likelihoods in simulation 
and perceptual modules in deployment.

\section{Evaluation of LEGS-POMDP}
We evaluate whether the LEGS-POMDP framework enables robust and efficient multimodal object search under uncertainty.
The evaluation proceeds in three stages: (i) modular tests of gesture and language grounding, (ii) gridworld simulations comparing solvers, modalities, and environment complexity, and (iii) real-robot deployment on the Boston Dynamics Spot. 
Across these settings, results show that multimodal grounding improves robustness, PO-UCT enhances planning reliability, and the integrated system achieves strong performance in both simulation and real-world.
\subsection{Modular Evaluation}
We evaluate gesture and language grounding to examine how each modality resolves referential ambiguity. Using the YouRefIt dataset~\cite{chen2021yourefitembodiedreferenceunderstanding}, which contains 4,221 annotated pointing and language instances, we benchmark each modality in isolation. The results highlight complementary strengths and limitations, motivating the integration in LEGS-POMDP for robust multimodal grounding.

\subsubsection{\textbf{Gesture Grounding}} 
Formalized gesture representations provides structured likelihoods that can be directly incorporated into a POMDP observation model, yielding both probabilistic reasoning and interpretability for downstream belief updates. We evaluate gesture grounding to test whether different pointing representations can robustly capture human intent and identify which representation provides the most reliable basis for integration into a decision-making framework. 
We hypothesize that using a probabilistic cone representation with cues from multiple skeletal landmarks yields improved robustness and accuracy compared to single-vector baselines, especially in the presence of pose estimation noise.

\begin{figure}[t]
  \centering
  \includegraphics[width=\linewidth]{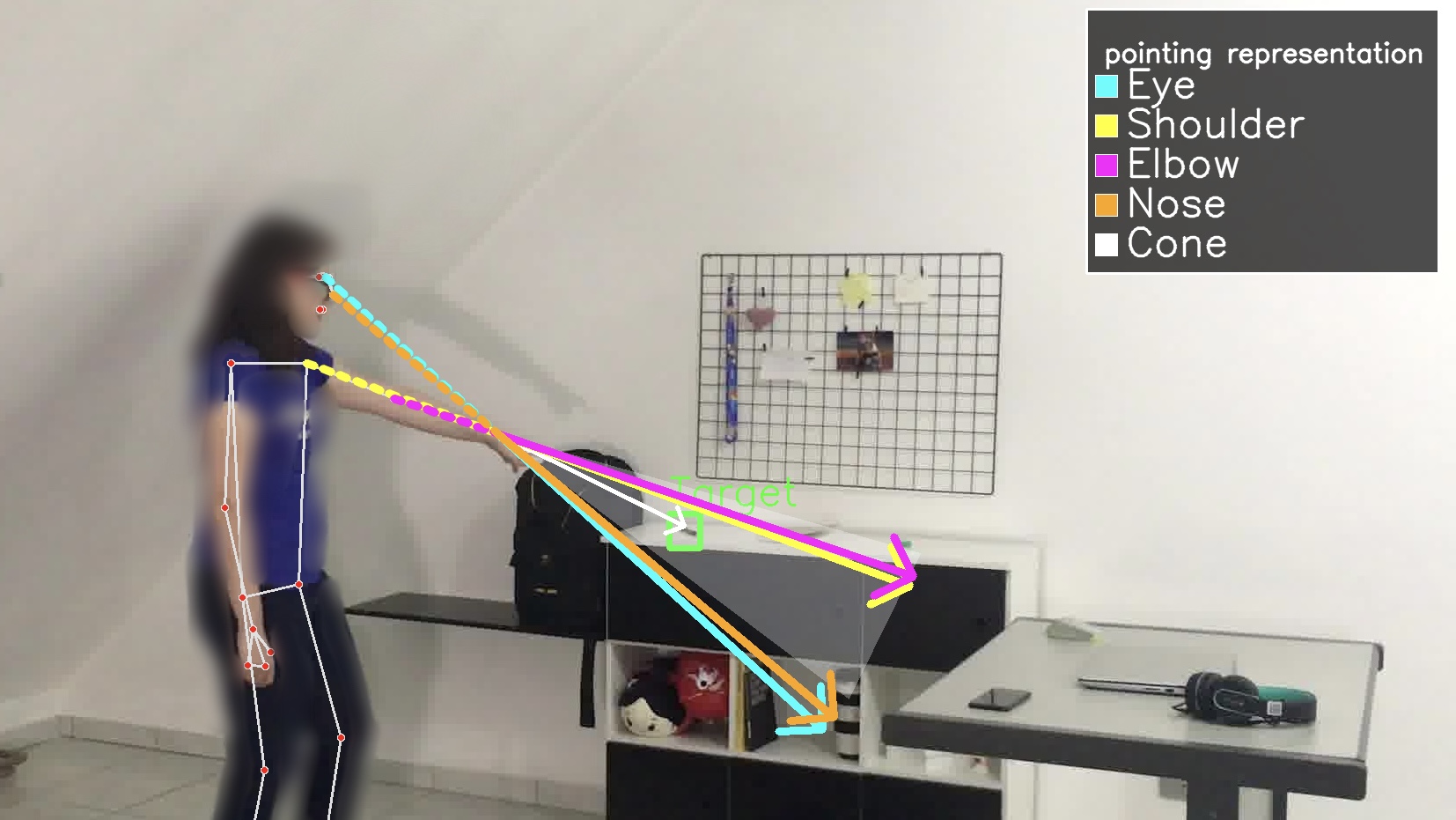}
  \caption{Example frame showing different vector- and cone-based models of the pointing direction, with the target marked in green.}
  \Description{Example frame showing different vector- and cone-based models of the pointing direction, with the target marked in green.}
  \label{fig:gesture_evaluation_demo}
\end{figure}

We compare four body landmark vectors (eye-to-wrist, nose-to-wrist, shoulder-to-wrist, and elbow-to-wrist) with a gesture cone representation that merges vector cues to form a probabilistic region of reference. All pointing vectors are anchored at the wrist, which provides a stable and consistently detectable landmark in dynamic scenes. While finer hand landmarks could in principle yield more precise estimates, hand detection is often less reliable under occlusion or motion, making the wrist a more robust endpoint for downstream analysis. MediaPipe~\cite{lugaresi_mediapipe_2019} is used for skeleton detection, achieving a 92.6\% human detection rate on the YouRefIt dataset; all evaluation is conditioned on detected frames. Fig.~\ref{fig:gesture_evaluation_demo} shows an example frame with different vector- and cone-based pointing representations, with the target object highlighted in green.

Performance is evaluated using two metrics: (i) \textit{Coverage Accuracy @25\%} is defined as the percentage of samples where the predicted cone overlaps more than 25\% of the ground-truth bounding box area. For single-vector representations, we used a $15^\circ$ fixed opening angle, while for the gesture cone the opening angle is dynamically determined from the spread of included vectors. (ii) \textit{Angular Error} ($\theta_{\text{diff}}$), the deviation in degrees between the predicted pointing direction and the ground-truth reference. 


\begin{table}[t]
\vspace{-2mm}
\centering
\caption{Comparison of gesture representation for pointing estimation for pointing estimation. 
Metrics are reported as mean $\pm$ 95\% confidence interval (CI).}
\label{tab:gesture_representation}
\begin{tabular}{lccc}
\hline
\textbf{Pointing Representation} & \textbf{Cov. @25\%} & \textbf{$\theta_{\text{diff}}$ ($^\circ$)}  \\
\hline
Eye-to-Wrist       & 0.718 $\pm$ 0.014 & 24.4 $\pm$ 0.8  \\
Nose-to-Wrist      & 0.746 $\pm$ 0.014 & 23.2 $\pm$ 0.8  \\
Shoulder-to-Wrist  & 0.865 $\pm$ 0.011 & 17.0 $\pm$ 0.7  \\
Elbow-to-Wrist     & 0.772 $\pm$ 0.013 & 20.2 $\pm$ 0.8  \\
Gesture Cone       & \textbf{0.890 $\pm$ 0.010} & \textbf{14.4 $\pm$ 0.4}  \\
\hline
\end{tabular}
\vspace{-2mm}
\end{table}


Results in Table~\ref{tab:gesture_representation} show that the gesture cone achieves the lowest angular error ($14.4^\circ$) and the highest coverage accuracy (0.89). Compared to the best single-vector baseline (shoulder-to-wrist, 0.865), it improves coverage by $2.5\%$ and reduces angular error by $2.6^\circ$. In contrast, single-vector representations are more sensitive to target height and arm posture. We also observed qualitative differences across pointing representations. For low-lying targets, the elbow-to-wrist vector was often the most accurate, but it tended to overshoot for elevated targets. Wrist flexion or extension also altered pointing reliability. In cluttered scenes, the nose-to-wrist vector sometimes provided better disambiguation. Moreover, under frontal-facing condition, gaze-based vectors occasionally diverged from arm-based vectors, making generalization with single vector representation difficult. By averaging cues, the gesture cone stabilizes performance across varied conditions, making it a more reliable representation for downstream POMDP belief updates.

\subsubsection{\textbf{Visual Grounding}} 
Robust language grounding is essential for downstream decision-making in the POMDP framework since grounding failure leads to corrupted belief state, which in turn biases future planning toward incorrect targets. Thus, we evaluate visual grounding to test how well different grounding strategies resolve referential expressions.  
Our hypothesis is that decoupling perception and reasoning via a two-stage SoM pipeline (segmentation + LLM classification) yields more accurate and interpretable grounding than end-to-end detector-based methods.

\begin{figure}[t]
  \centering
  \includegraphics[width=\linewidth]{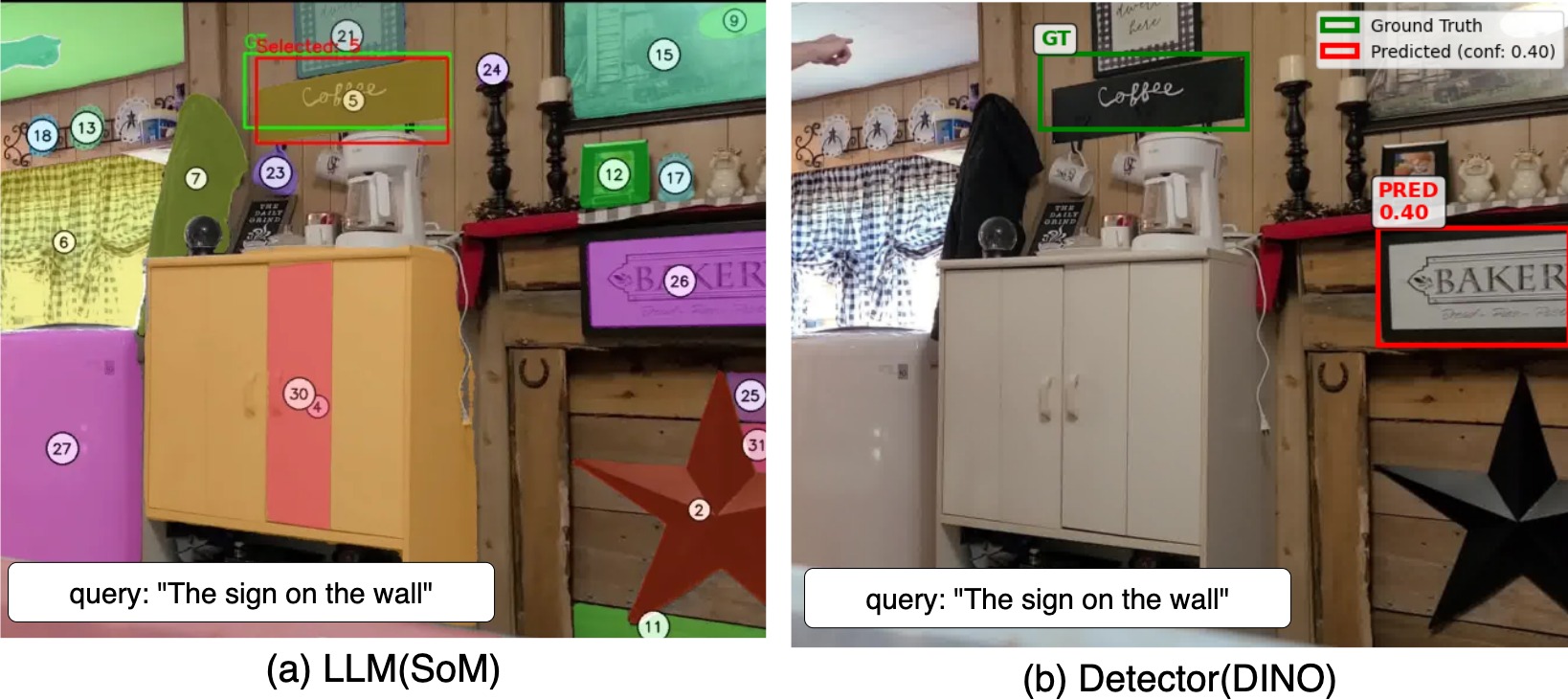}
  \caption{Visual grounding comparison between SoM prompting and a detector baseline (GroundingDINO).}
  \Description{Panels comparing detections from a VLM against SoM; SoM isolates intended objects more accurately when ambiguity is present.}
  \label{fig:languageTable}
\end{figure}

We conduct a comparative analysis between two distinct visual grounding paradigms: a detector-based baseline using GroundingDINO and an LLM-based Set-of-Marks (SoM) approach that integrates SAM2 segmentation with GPT-4o reasoning capabilities, as shown in Fig.\ref{fig:languageTable}. Queries may involve object attributes and/or spatial relations, allowing us to evaluate grounding robustness across different linguistic conditions. 
Performance is evaluated using three complementary metrics: 
(i) \textit{Detection Accuracy} (det. Acc) measures whether the system successfully localizes any candidate region for the queried object, providing a recall-oriented view of localization;
(ii) \textit{$IoU@25\%$}; (iii) \textit{Grounding Accuracy} (Grounding Acc) evaluates whether the predicted region correctly overlaps the GT target, reflecting semantic correctness of the grounding.

The Set-of-Marks (SoM) approach achieved higher detection success (92.3\% vs.\ 87.8\%) and grounding accuracy (91.4\% vs.\ 62.4\%) compared to the detector-based baseline. However, its IoU@25\% was lower (0.219 vs.\ 0.501), largely because SAM2 produced fine-grained masks smaller than the annotated bounding boxes, leading to underestimated overlap. This highlights that SoM is limited by its dependence on segmentation quality. In addition, SoM incurs significantly higher inference time due to its two-stage pipeline, trading efficiency for robustness in resolving referring expressions.


\begin{table}[t]
\centering
\caption{Grounding Success rate under different language conditions. 
LLM grounding (SoM: SAM2 + GPT-4o) vs. detector grounding (GroundingDINO).}
\begin{tabular}{lcc}
\toprule
\textbf{Grounding Acc}  & \textbf{LLM (SoM)} & \textbf{Detector (DINO)}\\
\midrule
None                & 0.793 & 0.603 \\
Spatial             & 0.957 & 0.577 \\
Attribute           & 0.944 & 0.665 \\
Spatial + Attribute & 0.818 & 0.956 \\
\bottomrule
\end{tabular}
\label{tab:language_conditions}
\vspace{-3mm}
\end{table}

Results in Table~\ref{tab:language_conditions} report grounding accuracy conditioned on successful detection. We observe that the detector-based baseline struggles in most cases: when only spatial (0.577) or attribute (0.665) references are provided, accuracy drops sharply in cluttered scenes. As illustrated in Fig.~\ref{fig:languageTable}, the SoM approach (a) correctly grounds the query “the sign on the wall” by isolating the intended region. In contrast, the detector baseline (b) detects a sign but misinterprets the spatial qualifier “on the wall,” incorrectly selecting a lower sign. The detector performs relatively well when both spatial and attribute cues are combined (0.956), suggesting that it leverages explicitly learned patterns from training data when richer descriptions are available. In contrast, the SoM approach maintains consistently high performance across single-reference conditions, achieving 0.957 on spatial and 0.944 on attribute queries. However, SoM accuracy decreases with compounded descriptions (0.818), likely due to segmentation ambiguities and language model parsing errors. The results indicate that LLM-based SoM grounding generalizes more robustly to underrepresented linguistic conditions, while the detector benefits more from detailed but less natural multi-cue descriptions. Importantly, SoM’s robustness in handling single but ambiguous references provides more stable observation likelihoods, reducing the risk of belief corruption in downstream POMDP planning.

\subsection{System Evaluation}
We evaluate our system in a grid-world simulation environment designed to capture the challenges of multimodal instruction following. The environment consists of grid cells populated with target objects, distractors, and static obstacles, requiring the agent to actively explore while maintaining a belief state over possible target locations. 
We prepared three grid environments of increasing spatial complexity (5×5, 10×10, 20×20).
Human inputs (gesture, language, or both) are injected as observations that directly influence belief updates, while distractors and obstacles introduce ambiguity and navigation costs. This setup allows us to systematically vary environment size and ambiguity, and to test how different modalities and solvers affect success, efficiency, and belief convergence.




\subsubsection{\textbf{Solver Comparison}} 
Solver comparison test determines whether sophisticated planning algorithms with principled belief representations enhance robustness in ambiguous visual grounding tasks compared to simple approaches. 
Four solvers are compared under a consistent observation model and reward function. 
The \textit{Greedy} baseline always executes the \texttt{Find} action as soon as any object is observed, ignoring uncertainty and planning. 
The \textit{Belief Heuristic} policy moves toward the grid cell with the highest current belief, considering only the most likely target location at each step.
\textit{POMCP} is a Monte Carlo Tree Search–based solver that leverages a particle belief representation for scalable online planning. 
Finally, \textit{PO-UCT} extends the UCT algorithm with deeper lookahead, balancing exploration and exploitation to improve planning under uncertainty. 
We hypothesize that principled POMDP solvers (POMCP and PO-UCT) will demonstrate more robust performance compared to heuristic baselines (Greedy and Belief Heuristic).

We conducted controlled experiments with no human input to isolate planning performance from perceptual challenges. Approximately 100 independent trials per solver-belief representation configuration is executed to ensure statistical reliability, with random initialization of object locations and agent starting positions.
Performance was evaluated using three metrics: (i) \textit{Success rate}, the fraction of trials in which the agent correctly identified and executed a \texttt{Find} action on the target object; (ii) \textit{Total steps}, mean number of actions required until task completion, measuring exploration effectiveness; and (iii) \textit{Total time}, total execution time including both planning overhead and action execution.



\begin{table}[t]
\centering
\caption{Solver performance under histogram vs. particle belief. Metrics are mean $\pm$ 95\% CI over all trials.}
\resizebox{\columnwidth}{!}{
\begin{tabular}{lcccc}
\toprule
\textbf{Belief} & \textbf{Solver} & \textbf{Success} & \textbf{Steps} & \textbf{Time [s]} \\
\midrule
\textbf{Histogram} & Heuristic & 0.68 $\pm$ 0.11 & 111.3 $\pm$ 24.0 & 12.0 $\pm$ 2.6 \\
                  & Greedy    & 0.63 $\pm$ 0.11 & 227.7 $\pm$ 20.4 & 24.6 $\pm$ 2.2 \\
                  & PO-UCT    & 0.96 $\pm$ 0.06 & 124.9 $\pm$ 14.7 & 32.2 $\pm$ 8.3 \\
\midrule
\textbf{Particles} & Heuristic & 0.21 $\pm$ 0.10 & 42.3 $\pm$ 7.9   & 4.7 $\pm$ 0.9  \\
                  & Greedy    & 0.27 $\pm$ 0.10 & 183.6 $\pm$ 18.4 & 20.5 $\pm$ 2.1 \\
                  & POMCP     & 0.24 $\pm$ 0.09 & 183.4 $\pm$ 20.2 & 36.8 $\pm$ 4.5 \\
                  & PO-UCT    & 0.45 $\pm$ 0.11 & 121.4 $\pm$ 10.3 & 30.6 $\pm$ 6.3 \\
\bottomrule
\end{tabular}
}
\vspace{-3mm}
\label{tab:solver_comparison}
\end{table}
Table ~\ref{tab:solver_comparison} demonstrates that PO-UCT achieves optimal performance under histogram belief representation, achieving $96\%$ success rate while maintaining competitive step counts and reasonable computational overhead. The Greedy baseline exhibits poor success rates despite minimal planning time, while POMCP shows inconsistent performance with higher variance in both success and efficiency metrics.
Under particle belief representation, all solvers experience degraded performance due to increased representational noise and approximation errors in belief updates. However, PO-UCT maintains the smallest performance degradation, suggesting greater robustness to belief representation quality.
The results validate our hypothesis that planning depth and stable belief representation are essential for reliable performance in ambiguous visual grounding tasks. While heuristic approaches offer computational efficiency, they sacrifice reliability. PO-UCT emerges as the optimal balance between robustness and efficiency.

\subsubsection{\textbf{Modality Evaluation}}
Human inputs directly influence the POMDP observation model; grounding failures or missing modalities can therefore corrupt belief updates. We evaluate how individual modalities guide exploration and how multimodal fusion improves robustness under ambiguity.
We fix the solver to PO-UCT and evaluate across five environments (small, small-ambiguous, medium, large, and large-ambiguous) under seven instruction conditions: no input, gesture-only, language-only, multimodal, wrong gesture, wrong language, and conflicted multimodal input. Performance is measured by success rate, steps to completion, and total time, with belief dynamics analyzed via max-belief and target-belief convergence over 10 trials per condition.





\begin{table}[t]
\centering
\caption{Modality comparison. Success rate, steps, and time are reported with 95\% confidence intervals.}
\label{tab:modality_comparison}
\resizebox{\columnwidth}{!}{
\begin{tabular}{lccc}
\hline
\textbf{Modality} & \textbf{Success Rate} & \textbf{Steps} & \textbf{Time [s]} \\
\hline
multimodal conflicted             & $0.024 \pm 0.007$ & $59.0 \pm 17.3$ & $21.1 \pm 12.8$ \\
wrong language         & $0.093 \pm 0.064$ & $95.9 \pm 35.2$ & $24.5 \pm 10.8$ \\
wrong gesture          & $0.170 \pm 0.049$ & $91.2 \pm 33.1$ & $22.8 \pm 11.1$ \\
No Input    & $0.482 \pm 0.130$ & $162.3 \pm 35.2$ & $37.0 \pm 10.4$ \\
Gesture     & $0.618 \pm 0.045$ & $122.5 \pm 25.5$ & $23.2 \pm 4.7$  \\
Language    & $0.710 \pm 0.057$ & $95.8 \pm 34.4$  & $20.4 \pm 5.4$  \\
Multimodal  & $0.888 \pm 0.073$ & $76.8 \pm 27.4$  & $16.7 \pm 5.6$  \\
\hline
\end{tabular}
}
\vspace{-3mm}
\end{table}


As shown in Table~\ref{tab:modality_comparison}, multimodal input achieves the highest success rate ($0.888 \pm 0.073$) with the fewest steps and fastest completion time, validating the complementarity of gesture and language.  While this result is not surprising, it demonstrates that our POMDP-based approach and solver are capable of successfully fusing information across multiple modalities to achieve improved performance.
Language-only ($0.710 \pm 0.057$) and gesture-only ($0.618 \pm 0.045$) perform moderately, while no-instruction drops sharply to $0.482 \pm 0.130$. 
Time usage shows a similar trend, with multimodal trials completing in $16.7$ seconds on average, nearly half of the no-instruction condition. 
These results highlight the complementary effect of gesture and language, showing that combining modalities improves task success and efficiency. 
In contrast, wrong gesture ($0.170 \pm 0.049$), wrong language ($0.093 \pm 0.064$), and especially conflicted multimodal input ($0.024 \pm 0.007$) nearly always fail, highlighting how erroneous inputs corrupt the belief state. 
\begin{figure}[t]
  \centering
  \includegraphics[width=\linewidth]{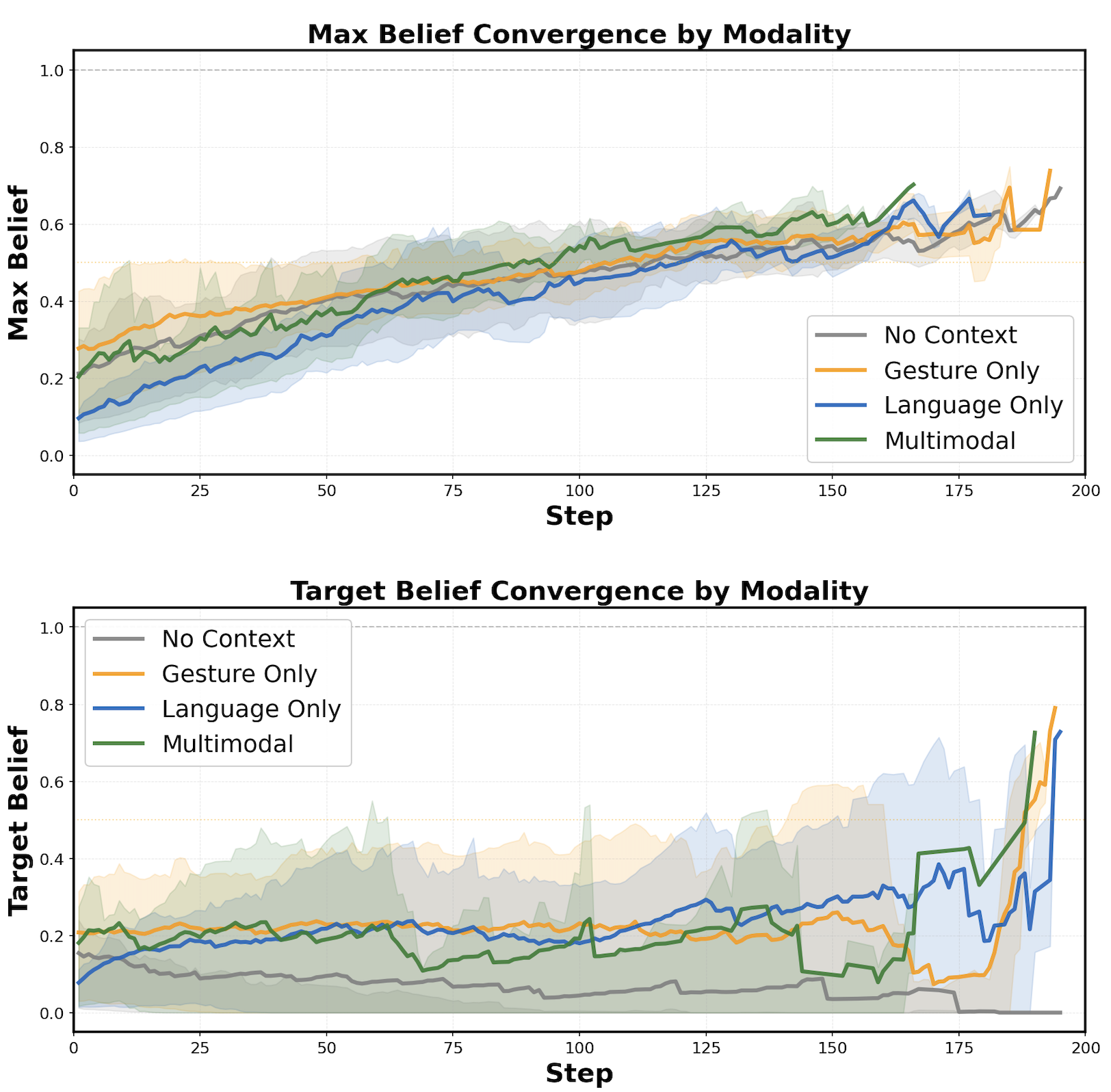}
  \caption{Belief convergence in the \textbf{large environment}. (Top) Max-belief traces show how certainty in the most likely state evolves over time. 
  (Bottom) Target-belief traces show probability mass assigned to the true target. }
  \Description{Belief convergence comparison}
  \label{fig:belief_convergency}
\end{figure}

In the large environment, belief convergence further illustrates the advantages of multimodal input. 
As shown in Fig.~\ref{fig:belief_convergency}, max-belief curves across modalities grow at similar rates, but multimodal trials terminate in fewer steps, reflecting faster convergence and decision-making efficiency. 
Target-belief curves reveal a sharper contrast: without instruction, belief quickly collapses toward distractors, while any valid human input yields sustained growth in target belief. These findings confirm that multimodal guidance mitigates the challenges of large, ambiguous environments by stabilizing belief updates and accelerating task completion.

\begin{figure}[t]
  \centering
  \includegraphics[width=\linewidth]{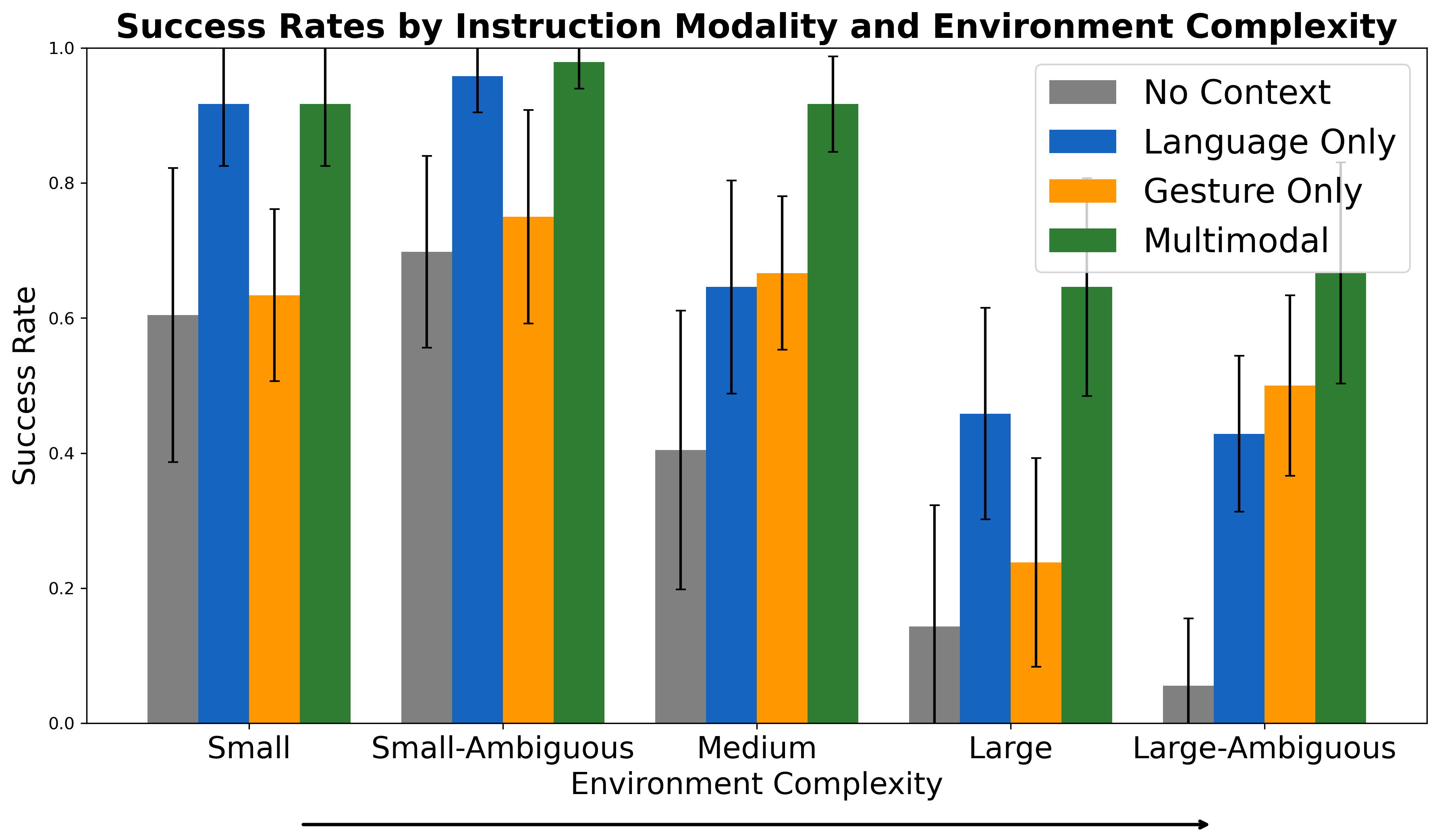}
  \caption{Success rates by instruction modality across environments of increasing complexity.}
  \Description{Success rates by instruction modality across environments of increasing complexity.}
  \label{fig:env_complexity}
\end{figure}

Figure~\ref{fig:env_complexity} shows how environmental complexity directly impacts grounding performance. 
As the environment becomes larger and more ambiguous, the performance of single-modality instructions degrades sharply. The no-instruction baseline collapses almost entirely in these cases. In contrast, multimodal input maintains relatively high success even under severe ambiguity, demonstrating that combining gesture and language provides robustness against increasing environmental complexity. 
By jointly analyzing solvers, modalities, and environment complexity, we establish that multimodal POMDP planning is both more robust and more efficient in ambiguous search settings.  This also demonstrates where an end-to-end approach trained on datasets would fall short in the large ambiguous environment because it would not be able to systematically search.  In the future, we plan to investigate end-to-end models with memory that can generalize in this way. 

\begin{figure*}[t]
  \centering
  \includegraphics[width=\textwidth]{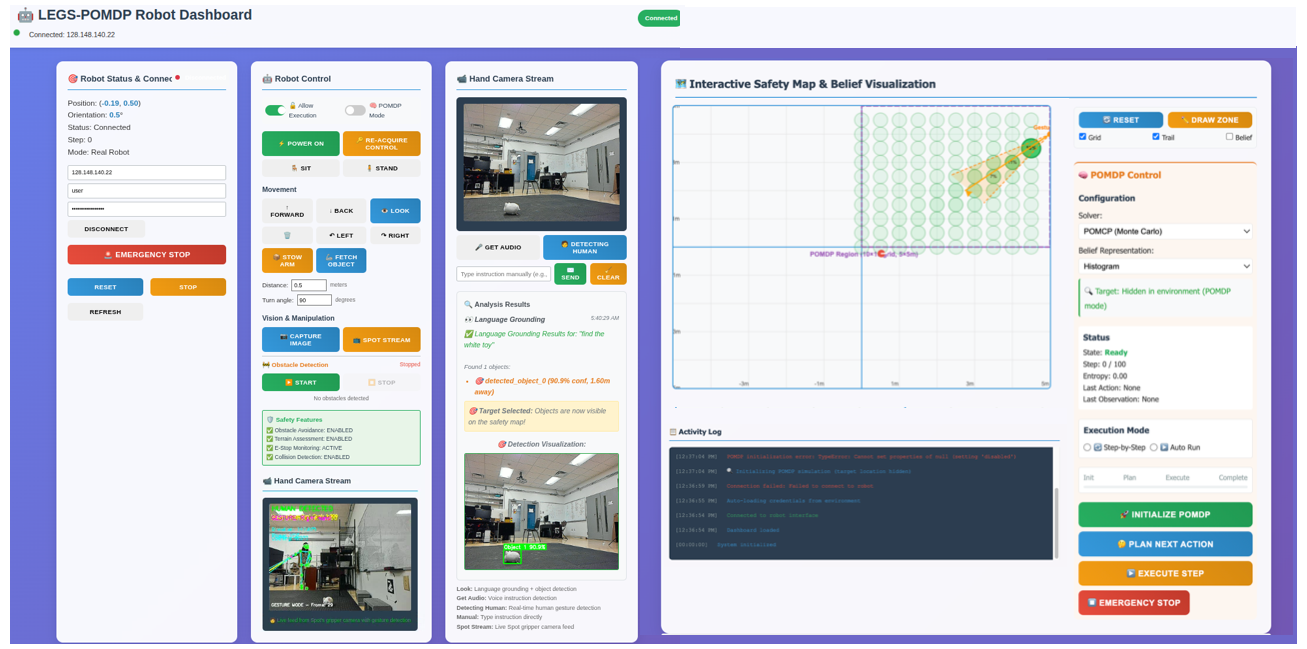}
  \caption{LEGS-POMDP robot testing and dashboard interface. The figure illustrates both the real-robot experimental setup and the integrated UI, which visualizes multimodal grounding, POMDP belief states, and robot control in real time.}
  \Description{LEGS-POMDP robot testing and dashboard interface. The figure illustrates both the real-robot experimental setup and the integrated UI, which visualizes multimodal grounding, POMDP belief states, and robot control in real time.}
  \label{fig:robot demo figure}
\end{figure*}

\begin{figure}
  \centering
  \includegraphics[width=\linewidth]{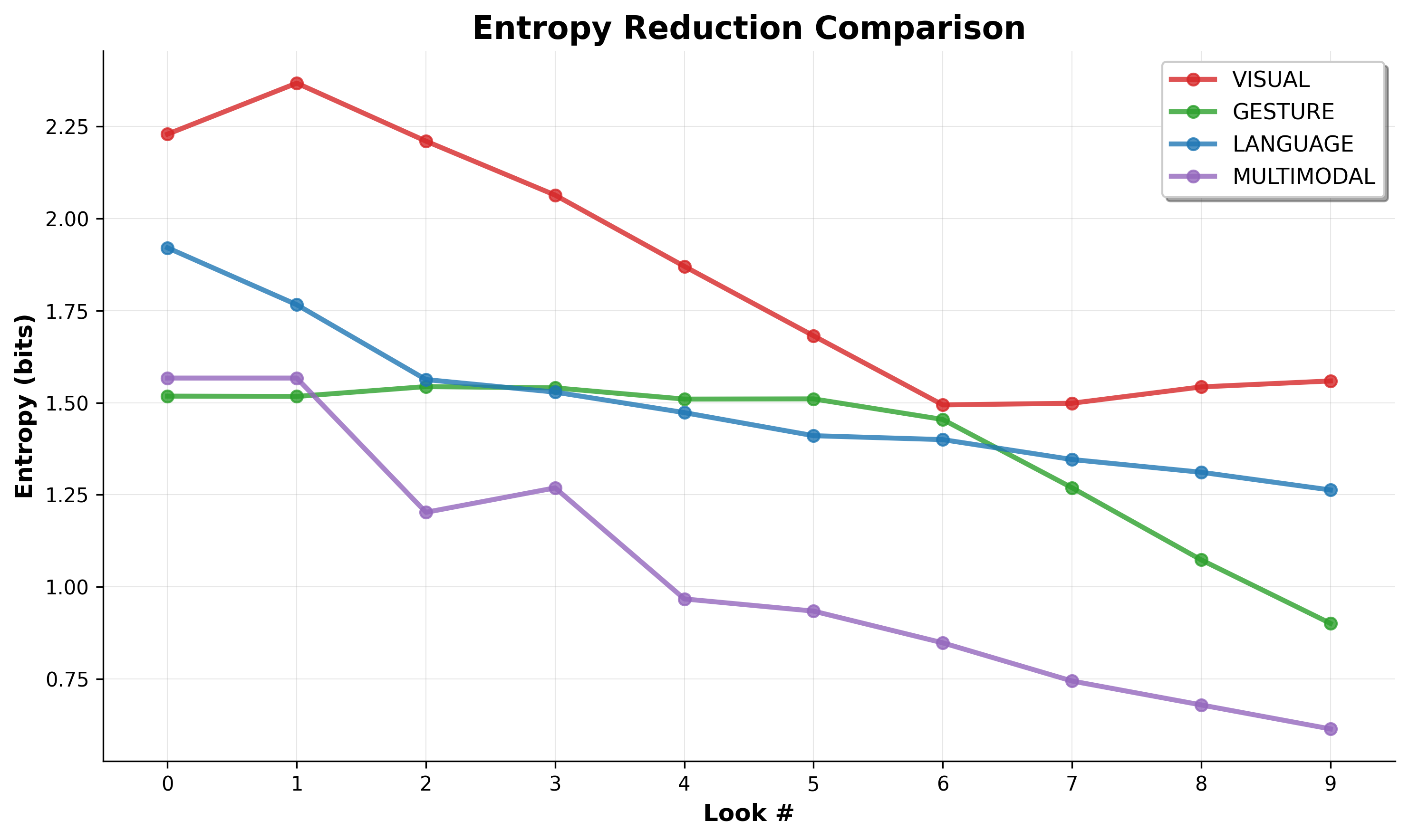}
  \caption{Entropy loss curve.}
  \Description{LEGS-POMDP robot testing and dashboard interface. The figure illustrates both the real-robot experimental setup and the integrated UI, which visualizes multimodal grounding, POMDP belief states, and robot control in real time.}
  \label{fig:entropy_reduction}
\end{figure}

\subsection{Robot Testing}
The goal of this evaluation is to test whether the LEGS-POMDP framework transfers from simulation to a real-world platform,  and how different modalities influence belief uncertainty under realistic conditions. One advantage of our modular approach is the ability to easily transfer between different robot hardware and different environments without collecting additional data or retraining, since each module is already trained on internet-scale data.

To evaluate whether multimodal input accelerates disambiguation, we conducted an ablation study in a 10×10 grid world with five objects, including three identical red cups placed in different locations to induce ambiguity. Without allowing the robot to execute move actions, we measured how belief uncertainty changed over 10 observation steps($A_\text{LOOK}$). As shown in Fig.~\ref{fig:entropy_reduction}, the multimodal (G+L) condition achieves the steepest entropy reduction rate, reducing entropy by $60.8\%$. Gesture alone also contributes strongly, achieving a $40.6\%$ reduction, while unimodal visual and language conditions reduce entropy by $30.1\%$ and $34.2\%$, respectively. These results indicate that both gesture and multimodal inputs more effectively narrow the prior belief compared to language and vision baselines. We further validated this trend by demonstrating successful execution of the object search task on the robot.


\section{Conclusion and Future Work}
We presented LEGS-POMDP, a multimodal POMDP framework for gesture- and language-conditioned object search under uncertainty. Simulation and modular evaluations show that multimodal fusion consistently outperforms single-modality baselines with an average success rate of 89\% in challenging simulated environments.
On a real quadruped mobile manipulator, we demonstrated the feasibility of the proposed framework by qualitatively validating multimodal grounding and uncertainty reduction in physical settings.
Multimodal fusion substantially improves disambiguation in human-instructed object search, particularly under high angular ambiguity. 
Cone-based gesture likelihoods capture spatial intent, while SoM-based language grounding provides semantic specificity. 
Weighted log-likelihood fusion enables more robust belief updates and faster convergence than unimodal alternatives.

Several limitations remain. 
Our fusion model assumes conditional independence between modalities, simplifying belief updates but ignoring potential correlations such as alignment between deictic language and pointing gestures. 
Our system also relies on accurate visual segmentation, where errors may degrade perception and downstream belief updates in cluttered or dynamic environments. 
While we demonstrate feasibility in simulation and on a real robot, the scale and diversity of real-world experiments remain limited.

Future work will explore richer multimodal integration, including tactile input and additional gesture types such as iconic gestures, as well as user studies in naturalistic environments to better understand how non-expert users employ multimodal communication during collaborative object search.
These directions aim to support more natural, robust, and adaptive human–robot interaction in open-world settings.

\begin{acks}
We thank Daphna Buchsbaum, Madeline Pelgrim, Jeff Huang, and Erin Hedlund-Botti for discussions and ideas adjacent to this work.
This work was supported in part by the National Science Foundation under Award No.~2433429 through the AI Research Institute on Interaction for AI Assistants (ARIA) and the Long-Term Autonomy for Ground and Aquatic Robotics program (Grant No.~GR5250131), and by the Office of Naval Research under Agreement No.~N00014-24-1-2784 and the ONR MURI program under Grant No.~N00014-24-1-2603.
\end{acks}
\bibliographystyle{ACM-Reference-Format}
\bibliography{bib/LEGS.bib}

@inproceedings{cohen2024ground,
  title={A Survey of Robotic Language Grounding: Tradeoffs between Symbols and Embeddings},
  author={Cohen, Vanya and Liu, Jason Xinyu and Mooney, Raymond and Tellex, Stefanie and Watkins, David},
  booktitle={International Joint Conference on Artificial Intelligence (IJCAI)},
  year={2024},
}

@inproceedings{liu2023lang2ltl,
    title={Grounding Complex Natural Language Commands for Temporal Tasks in Unseen Environments},
    author={Jason Xinyu Liu and Ziyi Yang and Ifrah Idrees and Sam Liang and Benjamin Schornstein and Stefanie Tellex and Ankit Shah},
    booktitle={Conference on Robot Learning (CoRL)},
    year={2023},
}

@inproceedings{liu2024lang2ltl2,
  title={{Lang2LTL}-2: Grounding Spatiotemporal Navigation Commands Using Large Language and Vision-Language Models},
  author={Liu, Jason Xinyu and Shah, Ankit and Konidaris, George and Tellex, Stefanie and Paulius, David},
  booktitle={IEEE/RSJ International Conference on Intelligent Robots and Systems (IROS)},
  year={2024}
}

@inproceedings{hsiung2022generalizing,
  title={Generalizing to new domains by mapping natural language to lifted LTL},
  author={Hsiung, Eric and Mehta, Hiloni and Chu, Junchi and Liu, Xinyu and Patel, Roma and Tellex, Stefanie and Konidaris, George},
  booktitle={IEEE International Conference on Robotics and Automation},
  year={2022}
}

@inproceedings{liang2022cap,
    title={Code as Policies: Language Model Programs for Embodied Control},
    author={Liang, Jacky and Wenlong Huang and Fei Xia and Peng Xu and Karol Hausman and Brian Ichter and Pete Florence and Andy Zeng},
    booktitle={IEEE International Conference on Robotics and Automation},
    year={2023}
}

@inproceedings{octo2023,
    title={Octo: An Open-Source Generalist Robot Policy},
    author = {{Octo Model Team} and Dibya Ghosh and Homer Walke and Karl Pertsch and Kevin Black and Oier Mees and Sudeep Dasari and Joey Hejna and Charles Xu and Jianlan Luo and Tobias Kreiman and {You Liang} Tan and Dorsa Sadigh and Chelsea Finn and Sergey Levine},
    booktitle={Robotics: Science and Systems},
    year = {2024},
}

@inproceedings{brohan2023rt1,
  title = {{RT}-1: Robotics Transformer for Real-World Control at Scale},
  author = {Anthony Brohan and Noah Brown and others},
  booktitle = {Robotics: Science and Systems},
  year = 2023
}

@inproceedings{zitkovich2023rt2,
  title={{RT}-2: Vision-Language-Action Models Transfer Web Knowledge to Robotic Control},
  author={Anthony Brohan and Noah Brown and others},
  booktitle={Conference on Robot Learning},
  year={2023}
}

@inproceedings{padalkar2024rtx,
  title={Open {X-E}mbodiment: Robotic Learning Datasets and {RT-X} Models},
  author={Abby O'Neill and Abdul Rehman and others},
  booktitle={IEEE International Conference on Robotics and Automation},
  year={2024}
}

@article{kim2024openvla,
  title={Openvla: An open-source vision-language-action model},
  author={Kim, Moo Jin and Pertsch, Karl and Karamcheti, Siddharth and Xiao, Ted and Balakrishna, Ashwin and Nair, Suraj and Rafailov, Rafael and Foster, Ethan and Lam, Grace and Sanketi, Pannag and others},
  journal={arXiv preprint arXiv:2406.09246},
  year={2024}
}

@article{black2024pi0,
  title={$\pi$0: A vision-language-action flow model for general robot control. CoRR, abs/2410.24164, 2024. doi: 10.48550},
  author={Black, Kevin and Brown, Noah and Driess, Danny and Esmail, Adnan and Equi, Michael and Finn, Chelsea and Fusai, Niccolo and Groom, Lachy and Hausman, Karol and Ichter, Brian and others},
  journal={arXiv preprint ARXIV.2410.24164},
  year={2024}
}

@inproceedings{liu2024grounding,
  title={Grounding dino: Marrying dino with grounded pre-training for open-set object detection},
  author={Liu, Shilong and Zeng, Zhaoyang and Ren, Tianhe and Li, Feng and Zhang, Hao and Yang, Jie and Jiang, Qing and Li, Chunyuan and Yang, Jianwei and Su, Hang and others},
  booktitle={European conference on computer vision},
  pages={38--55},
  year={2024},
  organization={Springer}
}

@inproceedings{wandzel_multi-object_2019,
	title = {Multi-{Object} {Search} using {Object}-{Oriented} {POMDPs}},
	url = {https://ieeexplore.ieee.org/abstract/document/8793888},
	doi = {10.1109/ICRA.2019.8793888},
	urldate = {2025-03-13},
	booktitle = {2019 {International} {Conference} on {Robotics} and {Automation} ({ICRA})},
	author = {Wandzel, Arthur and Oh, Yoonseon and Fishman, Michael and Kumar, Nishanth and Wong, Lawson L.S. and Tellex, Stefanie},
	month = may,
	year = {2019},
	note = {ISSN: 2577-087X},
	pages = {7194--7200},
}

@misc{zheng_system_2023,
	title = {A {System} for {Generalized} {3D} {Multi}-{Object} {Search}},
	url = {http://arxiv.org/abs/2303.03178},
	doi = {10.48550/arXiv.2303.03178},
	urldate = {2025-03-13},
	publisher = {arXiv},
	author = {Zheng, Kaiyu and Paul, Anirudha and Tellex, Stefanie},
	month = apr,
	year = {2023},
	note = {arXiv:2303.03178 [cs]},
}

@inproceedings{nguyen_language-conditioned_2023,
	address = {Detroit, MI, USA},
	title = {Language-{Conditioned} {Observation} {Models} for {Visual} {Object} {Search}},
	copyright = {https://doi.org/10.15223/policy-029},
	isbn = {978-1-6654-9190-7},
	url = {https://ieeexplore.ieee.org/document/10341492/},
	doi = {10.1109/IROS55552.2023.10341492},
	language = {en},
	urldate = {2025-03-13},
	booktitle = {2023 {IEEE}/{RSJ} {International} {Conference} on {Intelligent} {Robots} and {Systems} ({IROS})},
	publisher = {IEEE},
	author = {Nguyen, Thao and Hrosinkov, Vladislav and Rosen, Eric and Tellex, Stefanie},
	month = oct,
	year = {2023},
	pages = {10894--10901},
}

@misc{lin_gesture-informed_2023,
	title = {Gesture-{Informed} {Robot} {Assistance} via {Foundation} {Models}},
	url = {http://arxiv.org/abs/2309.02721},
	doi = {10.48550/arXiv.2309.02721},
	urldate = {2025-03-13},
	publisher = {arXiv},
	author = {Lin, Li-Heng and Cui, Yuchen and Hao, Yilun and Xia, Fei and Sadigh, Dorsa},
	month = sep,
	year = {2023},
	note = {arXiv:2309.02721 [cs]},
}

@misc{wang_thisthat_2024,
	title = {This\&{That}: {Language}-{Gesture} {Controlled} {Video} {Generation} for {Robot} {Planning}},
	shorttitle = {This\&{That}},
	url = {http://arxiv.org/abs/2407.05530},
	doi = {10.48550/arXiv.2407.05530},
	urldate = {2025-03-13},
	publisher = {arXiv},
	author = {Wang, Boyang and Sridhar, Nikhil and Feng, Chao and Merwe, Mark Van der and Fishman, Adam and Fazeli, Nima and Park, Jeong Joon},
	month = jul,
	year = {2024},
	note = {arXiv:2407.05530 [cs]},
}

@incollection{gibet_deixis_2006,
	address = {Berlin, Heidelberg},
	title = {Deixis: {How} to {Determine} {Demonstrated} {Objects} {Using} a {Pointing} {Cone}},
	volume = {3881},
	isbn = {978-3-540-32624-3 978-3-540-32625-0},
	shorttitle = {Deixis},
	url = {http://link.springer.com/10.1007/11678816_34},
	language = {en},
	urldate = {2025-03-13},
	booktitle = {Gesture in {Human}-{Computer} {Interaction} and {Simulation}},
	publisher = {Springer Berlin Heidelberg},
	author = {Kranstedt, Alfred and Lücking, Andy and Pfeiffer, Thies and Rieser, Hannes and Wachsmuth, Ipke},
	editor = {Gibet, Sylvie and Courty, Nicolas and Kamp, Jean-François},
	year = {2006},
	doi = {10.1007/11678816_34},
	note = {Series Title: Lecture Notes in Computer Science},
	pages = {300--311},
}

@misc{kobzarev_gestllm_2025,
	title = {{GestLLM}: {Advanced} {Hand} {Gesture} {Interpretation} via {Large} {Language} {Models} for {Human}-{Robot} {Interaction}},
	shorttitle = {{GestLLM}},
	url = {http://arxiv.org/abs/2501.07295},
	doi = {10.48550/arXiv.2501.07295},
	urldate = {2025-03-13},
	publisher = {arXiv},
	author = {Kobzarev, Oleg and Lykov, Artem and Tsetserukou, Dzmitry},
	month = jan,
	year = {2025},
	note = {arXiv:2501.07295 [cs]},
}

@article{pelgrim_find_2024,
	title = {Find it like a dog: {Using} {Gesture} to {Improve} {Object} {Search}},
	volume = {46},
	shorttitle = {Find it like a dog},
	url = {https://escholarship.org/uc/item/0nk6w9fd},
	language = {en},
	number = {0},
	urldate = {2025-03-13},
	journal = {Proceedings of the Annual Meeting of the Cognitive Science Society},
	author = {Pelgrim, Madeline Helmer and He, Ivy Xiao and Lee, Kyle and Pabari, Falak and Tellex, Stefanie and Nguyen, Thao and Buchsbaum, Daphna},
	year = {2024},
}

@misc{yang_interactive_2022,
	title = {Interactive {Robotic} {Grasping} with {Attribute}-{Guided} {Disambiguation}},
	url = {http://arxiv.org/abs/2203.08037},
	doi = {10.48550/arXiv.2203.08037},
	urldate = {2025-03-13},
	publisher = {arXiv},
	author = {Yang, Yang and Lou, Xibai and Choi, Changhyun},
	month = mar,
	year = {2022},
	note = {arXiv:2203.08037 [cs]},
}

@misc{yang_set--mark_2023,
	title = {Set-of-{Mark} {Prompting} {Unleashes} {Extraordinary} {Visual} {Grounding} in {GPT}-{4V}},
	url = {http://arxiv.org/abs/2310.11441},
	doi = {10.48550/arXiv.2310.11441},
	urldate = {2025-03-13},
	publisher = {arXiv},
	author = {Yang, Jianwei and Zhang, Hao and Li, Feng and Zou, Xueyan and Li, Chunyuan and Gao, Jianfeng},
	month = nov,
	year = {2023},
	note = {arXiv:2310.11441 [cs]},
}

@misc{zou_segment_2023,
	title = {Segment {Everything} {Everywhere} {All} at {Once}},
	url = {http://arxiv.org/abs/2304.06718},
	doi = {10.48550/arXiv.2304.06718},
	urldate = {2025-03-13},
	publisher = {arXiv},
	author = {Zou, Xueyan and Yang, Jianwei and Zhang, Hao and Li, Feng and Li, Linjie and Wang, Jianfeng and Wang, Lijuan and Gao, Jianfeng and Lee, Yong Jae},
	month = jul,
	year = {2023},
	note = {arXiv:2304.06718 [cs]},
}

@misc{li_semantic-sam_2023,
	title = {Semantic-{SAM}: {Segment} and {Recognize} {Anything} at {Any} {Granularity}},
	shorttitle = {Semantic-{SAM}},
	url = {http://arxiv.org/abs/2307.04767},
	doi = {10.48550/arXiv.2307.04767},
	urldate = {2025-03-13},
	publisher = {arXiv},
	author = {Li, Feng and Zhang, Hao and Sun, Peize and Zou, Xueyan and Liu, Shilong and Yang, Jianwei and Li, Chunyuan and Zhang, Lei and Gao, Jianfeng},
	month = jul,
	year = {2023},
	note = {arXiv:2307.04767 [cs]},
}

@misc{lugaresi_mediapipe_2019,
	title = {{MediaPipe}: {A} {Framework} for {Building} {Perception} {Pipelines}},
	shorttitle = {{MediaPipe}},
	url = {http://arxiv.org/abs/1906.08172},
	doi = {10.48550/arXiv.1906.08172},
	urldate = {2025-03-13},
	publisher = {arXiv},
	author = {Lugaresi, Camillo and Tang, Jiuqiang and Nash, Hadon and McClanahan, Chris and Uboweja, Esha and Hays, Michael and Zhang, Fan and Chang, Chuo-Ling and Yong, Ming Guang and Lee, Juhyun and Chang, Wan-Teh and Hua, Wei and Georg, Manfred and Grundmann, Matthias},
	month = jun,
	year = {2019},
	note = {arXiv:1906.08172 [cs]},
}

@inproceedings{silver_monte-carlo_nodate,
author = {Silver, David and Veness, Joel},
title = {Monte-Carlo planning in large POMDPs},
year = {2010},
publisher = {Curran Associates Inc.},
address = {Red Hook, NY, USA},
booktitle = {Proceedings of the 24th International Conference on Neural Information Processing Systems - Volume 2},
pages = {2164–2172},
numpages = {9},
location = {Vancouver, British Columbia, Canada},
series = {NIPS'10}
}

@inproceedings{perlmutter_situated_2016,
	title = {Situated {Language} {Understanding} with {Human}-like and {Visualization}-{Based} {Transparency}},
	isbn = {978-0-9923747-2-3},
	url = {http://www.roboticsproceedings.org/rss12/p40.pdf},
	doi = {10.15607/RSS.2016.XII.040},
	language = {en},
	urldate = {2025-08-19},
	booktitle = {Robotics: {Science} and {Systems} {XII}},
	publisher = {Robotics: Science and Systems Foundation},
	author = {Perlmutter, Leah and Kernfeld, Eric and Cakmak, Maya},
	year = {2016},
}

@Article{electronics14122346,
AUTHOR = {Kukier, Tymon and Wróbel, Alicja and Sienkiewicz, Barbara and Klimecka, Julia and Gonzalez, Antonio Galiza Cerdeira and Gajewski, Paweł and Indurkhya, Bipin},
TITLE = {An Empirical Study on Pointing Gestures Used in Communication in Household Settings},
JOURNAL = {Electronics},
VOLUME = {14},
YEAR = {2025},
NUMBER = {12},
ARTICLE-NUMBER = {2346},
URL = {https://www.mdpi.com/2079-9292/14/12/2346},
ISSN = {2079-9292},
DOI = {10.3390/electronics14122346}
}

@inproceedings{
tanada_pointing_nodate,
title={Pointing Gesture Understanding via Visual Prompting and Visual Question Answering for Interactive Robot Navigation},
author={Kosei Tanada and Shigemichi Matsuzaki and Kazuhito Tanaka and Shintaro Nakaoka and Yuki Kondo and Yuto Mori},
booktitle={First Workshop on Vision-Language Models for Navigation and Manipulation at ICRA 2024},
year={2024},
url={https://openreview.net/forum?id=sJjwtGvK5D}
}

@article{
doi:10.1126/scirobotics.adf6991,
author = {Theophile Gervet  and Soumith Chintala  and Dhruv Batra  and Jitendra Malik  and Devendra Singh Chaplot },
title = {Navigating to objects in the real world},
journal = {Science Robotics},
volume = {8},
number = {79},
pages = {eadf6991},
year = {2023},
doi = {10.1126/scirobotics.adf6991},
URL = {https://www.science.org/doi/abs/10.1126/scirobotics.adf6991},
eprint = {https://www.science.org/doi/pdf/10.1126/scirobotics.adf6991}}

@inproceedings{mao2016generation,
  title={Generation and comprehension of unambiguous object descriptions},
  author={Mao, Junhua and Huang, Jonathan and Toshev, Alexander and Camburu, Oana and Yuille, Alan L and Murphy, Kevin},
  booktitle={Proceedings of the IEEE conference on computer vision and pattern recognition},
  pages={11--20},
  year={2016}
}

@article{tellex_robots_nodate,
title = "Robots That Use Language",
author = "Stefanie Tellex and Nakul Gopalan and Hadas Kress-Gazit and Cynthia Matuszek",
note = "Publisher Copyright: Copyright {\textcopyright} 2020 by Annual Reviews. All rights reserve.",
year = "2020",
month = may,
day = "3",
doi = "10.1146/annurev-control-101119-071628",
language = "English (US)",
volume = "3",
pages = "25--55",
journal = "Annual Review of Control, Robotics, and Autonomous Systems",
issn = "2573-5144",
publisher = "Annual Reviews Inc.",
}

@misc{shridhar_interactive_2018,
	title = {Interactive {Visual} {Grounding} of {Referring} {Expressions} for {Human}-{Robot} {Interaction}},
	url = {http://arxiv.org/abs/1806.03831},
	doi = {10.48550/arXiv.1806.03831},
	urldate = {2025-08-19},
	publisher = {arXiv},
	author = {Shridhar, Mohit and Hsu, David},
	month = jun,
	year = {2018},
	note = {arXiv:1806.03831 [cs]},
}

@inproceedings{whitney_reducing_2017,
	title = {Reducing errors in object-fetching interactions through social feedback},
	url = {https://ieeexplore.ieee.org/document/7989121},
	doi = {10.1109/ICRA.2017.7989121},
	urldate = {2025-08-19},
	booktitle = {2017 {IEEE} {International} {Conference} on {Robotics} and {Automation} ({ICRA})},
	author = {Whitney, David and Rosen, Eric and MacGlashan, James and Wong, Lawson L. S. and Tellex, Stefanie},
	month = may,
	year = {2017},
	pages = {1006--1013},
}

@misc{zheng_spatial_2021,
	title = {Spatial {Language} {Understanding} for {Object} {Search} in {Partially} {Observed} {City}-scale {Environments}},
	url = {http://arxiv.org/abs/2012.02705},
	doi = {10.48550/arXiv.2012.02705},
	urldate = {2025-08-19},
	publisher = {arXiv},
	author = {Zheng, Kaiyu and Bayazit, Deniz and Mathew, Rebecca and Pavlick, Ellie and Tellex, Stefanie},
	month = jul,
	year = {2021},
	note = {arXiv:2012.02705 [cs]},
}

@misc{mani_point_2022,
	title = {Point and {Ask}: {Incorporating} {Pointing} into {Visual} {Question} {Answering}},
	shorttitle = {Point and {Ask}},
	url = {http://arxiv.org/abs/2011.13681},
	doi = {10.48550/arXiv.2011.13681},
	urldate = {2025-08-19},
	publisher = {arXiv},
	author = {Mani, Arjun and Yoo, Nobline and Hinthorn, Will and Russakovsky, Olga},
	month = feb,
	year = {2022},
	note = {arXiv:2011.13681 [cs]},
}

@misc{zhang_invigorate_2024,
	title = {{INVIGORATE}: {Interactive} {Visual} {Grounding} and {Grasping} in {Clutter}},
	shorttitle = {{INVIGORATE}},
	url = {http://arxiv.org/abs/2108.11092},
	doi = {10.48550/arXiv.2108.11092},
	urldate = {2025-08-19},
	publisher = {arXiv},
	author = {Zhang, Hanbo and Lu, Yunfan and Yu, Cunjun and Hsu, David and Lan, Xuguang and Zheng, Nanning},
	month = jan,
	year = {2024},
	note = {arXiv:2108.11092 [cs]},
}

@misc{chen2021yourefitembodiedreferenceunderstanding,
      title={YouRefIt: Embodied Reference Understanding with Language and Gesture}, 
      author={Yixin Chen and Qing Li and Deqian Kong and Yik Lun Kei and Song-Chun Zhu and Tao Gao and Yixin Zhu and Siyuan Huang},
      year={2021},
      eprint={2109.03413},
      archivePrefix={arXiv},
      primaryClass={cs.CV},
      url={https://arxiv.org/abs/2109.03413}, 
}

@article{caesar_islam_2022,
	title = {CAESAR: An Embodied Simulator for Generating Multimodal Referring Expression Datasets},
	author = {Islam, Md. Mofijul and Mirzaiee, Reza and Gladstone, Alexi and Green, Haley N. and Iqbal, Tariq},
	journal = {Neural Information Processing Systems},
	year = {2022},
	litmapsId = {260750214}
}

@misc{mane2025ges3vig,
      title={Ges3ViG: Incorporating Pointing Gestures into Language-Based 3D Visual Grounding for Embodied Reference Understanding}, 
      author={Atharv Mahesh Mane and Dulanga Weerakoon and Vigneshwaran Subbaraju and Sougata Sen and Sanjay E. Sarma and Archan Misra},
      year={2025},
      eprint={2504.09623},
      archivePrefix={arXiv},
      primaryClass={cs.CV},
      url={https://arxiv.org/abs/2504.09623}, 
}

@misc{nakamura2023deepointvisualpointingrecognition,
      title={DeePoint: Visual Pointing Recognition and Direction Estimation}, 
      author={Shu Nakamura and Yasutomo Kawanishi and Shohei Nobuhara and Ko Nishino},
      year={2023},
      eprint={2304.06977},
      archivePrefix={arXiv},
      primaryClass={cs.CV},
      url={https://arxiv.org/abs/2304.06977}, 
}

@inproceedings{lin2020multi,
  title={Multi-Level Structure vs. End-to-End-Learning in High-Performance Tactile Robotic Manipulation},
  author={Lin, Hsiu-Chin and Iba, Soshi and Ramos, Fabio},
  booktitle={Conference on Robot Learning (CoRL)},
  pages={516},
  year={2020},
  organization={PMLR}
}

@misc{chaplot2020objectgoalnavigationusing,
      title={Object Goal Navigation using Goal-Oriented Semantic Exploration}, 
      author={Devendra Singh Chaplot and Dhiraj Gandhi and Abhinav Gupta and Ruslan Salakhutdinov},
      year={2020},
      eprint={2007.00643},
      archivePrefix={arXiv},
      primaryClass={cs.CV},
      url={https://arxiv.org/abs/2007.00643}, 
}

@inproceedings{yokoyama2024vlfm,
  title={Vlfm: Vision-language frontier maps for zero-shot semantic navigation},
  author={Yokoyama, Naoki and Ha, Sehoon and Batra, Dhruv and Wang, Jiuguang and Bucher, Bernadette},
  booktitle={2024 IEEE International Conference on Robotics and Automation (ICRA)},
  pages={42--48},
  year={2024},
  organization={IEEE}
}

@misc{garg2024robohopsegmentbasedtopologicalmap,
      title={RoboHop: Segment-based Topological Map Representation for Open-World Visual Navigation}, 
      author={Sourav Garg and Krishan Rana and Mehdi Hosseinzadeh and Lachlan Mares and Niko Sünderhauf and Feras Dayoub and Ian Reid},
      year={2024},
      eprint={2405.05792},
      archivePrefix={arXiv},
      primaryClass={cs.RO},
      url={https://arxiv.org/abs/2405.05792}, 
}

@article{hughes2024foundations,
  title={Foundations of spatial perception for robotics: Hierarchical representations and real-time systems},
  author={Hughes, Nathan and Chang, Yun and Hu, Siyi and Talak, Rajat and Abdulhai, Rumaia and Strader, Jared and Carlone, Luca},
  journal={The International Journal of Robotics Research},
  volume={43},
  number={10},
  pages={1457--1505},
  year={2024},
  publisher={SAGE Publications Sage UK: London, England}
}

@article{wang2024sim2real,
  title={Sim-to-Real Transfer via 3D Feature Fields for Vision-and-Language Navigation},
  author={Wang, Zihan and Yang, Xiangyu and Liang, Jiahao and Xu, Jing and Luo, Yuehu and Yang, Zhiming and Zhang, Haojian and Hu, Xiaoyu and Wang, Yandong},
  journal={arXiv preprint arXiv:2406.09798},
  year={2024}
}

@misc{fu2024mobilealohalearningbimanual,
      title={Mobile ALOHA: Learning Bimanual Mobile Manipulation with Low-Cost Whole-Body Teleoperation}, 
      author={Zipeng Fu and Tony Z. Zhao and Chelsea Finn},
      year={2024},
      eprint={2401.02117},
      archivePrefix={arXiv},
      primaryClass={cs.RO},
      url={https://arxiv.org/abs/2401.02117}, 
}

@inproceedings{chi2023diffusionpolicy,
  title={Diffusion Policy: Visuomotor Policy Learning via Action Diffusion},
  author={Chi, Cheng and Feng, Siyuan and Du, Yilun and Xu, Zhenjia and Cousineau, Eric and Burchfiel, Benjamin and Song, Shuran},
  booktitle={Proceedings of Robotics: Science and Systems (RSS)},
  year={2023}
}

@article{geng2023gapartnet,
  title={GAPartNet: Cross-Category Domain-Generalizable Object Perception and Manipulation via Generalizable and Actionable Parts},
  author={Geng, Haoran and Xu, Helin and Zhao, Chengyang and Xu, Chao and Yi, Li and Huang, Siyuan and Wang, He},
  journal={arXiv preprint arXiv:2303.04137v5},
  year={2024}
}

@article{nickel2007visual,
  title={Visual recognition of pointing gestures for human-robot interaction},
  author={Nickel, Kai and Stiefelhagen, Rainer},
  journal={Image and Vision Computing},
  volume={25},
  number={12},
  pages={1875--1884},
  year={2007},
  publisher={Elsevier}
}

@inproceedings{nickel2003pointing,
  title={Pointing gesture recognition based on {3D}-tracking of face, hands and head orientation},
  author={Nickel, Kai and Stiefelhagen, Rainer},
  booktitle={Proceedings of the 5th International Conference on Multimodal Interfaces},
  pages={140--146},
  year={2003},
  organization={ACM}
}

@article{filipek2025empirical,
  title={An Empirical Study on Pointing Gestures Used in Communication in Household Settings},
  author={Filipek, Bart{\l}omiej and Banach, Marcin and Fran{\'c}, Olga and Zguda, Jan and Belter, Dominik},
  journal={Electronics},
  volume={14},
  number={12},
  pages={2346},
  year={2025},
  publisher={MDPI}
}
\end{document}